\documentclass[conference]{IEEEtran}
\IEEEoverridecommandlockouts

\usepackage{cite}
\usepackage{amsmath,amssymb,amsfonts}
\usepackage{graphicx}
\usepackage{textcomp}
\graphicspath{{./figs/}} 
\def\BibTeX{{\rm B\kern-.05em{\sc i\kern-.025em b}\kern-.08em
    T\kern-.1667em\lower.7ex\hbox{E}\kern-.125emX}}

\usepackage{algorithm}
\usepackage[noend]{algpseudocode}

\usepackage{graphics} 
\usepackage{amsmath} 
\usepackage{amssymb}  
\usepackage{adjustbox}
\usepackage{multirow}
\usepackage{subcaption}
\usepackage{booktabs}
\usepackage{booktabs}
\usepackage{tabularx}

\title{\LARGE \bf
Spiking Neural Network Phase Encoding for Cognitive Computing
}

\author{\IEEEauthorblockN{Lei Zhang}
\IEEEauthorblockA{\textit{Faculty of Engineering and Applied Science, University of Regina} \\
\textit{Regina, Saskatchewan, Canada} \\
\textit{Email:lei.zhang@uregina.ca}}
}

\begin{document}

\maketitle
\thispagestyle{empty}
\pagestyle{empty}

\begin{abstract}
This paper presents a novel approach for signal reconstruction using Spiking Neural Networks (SNN) based on the principles of Cognitive Informatics and Cognitive Computing. The proposed SNN leverages the Discrete Fourier Transform (DFT) to represent and reconstruct arbitrary time series signals. By employing N spiking neurons, the SNN captures the frequency components of the input signal, with each neuron assigned a unique frequency. The relationship between the magnitude and phase of the spiking neurons and the DFT coefficients is explored, enabling the reconstruction of the original signal. Additionally, the paper discusses the encoding of impulse delays and the phase differences between adjacent frequency components. This research contributes to the field of signal processing and provides insights into the application of SNN for cognitive signal analysis and reconstruction.
\end{abstract}

\begin{IEEEkeywords}
cognitive informatics, cognitive computing, artificial intelligence (AI), spiking neural network (SNN), phase encoding
\end{IEEEkeywords}

\section{Introduction}

The field of cognitive informatics and cognitive computing has seen significant advancements in recent years, paving the way for exciting possibilities in various industries\cite{YXWang18}. The state of the art in cognitive informatics and cognitive computing demonstrates remarkable progress in replicating human cognitive abilities and leveraging them to create intelligent systems and technologies\cite{CHEN2016688}. These advancements hold great potential for transforming industries, improving human-computer interaction, and addressing complex problems in various domains. 

\subsection{Cognitive Informatics}
Cognitive informatics focuses on the study of cognitive processes. Cognitive computing refers to systems that mimic human cognitive abilities, such as perception, reasoning, learning, and problem-solving. These systems leverage artificial intelligence (AI) techniques and cognitive models to process and analyze complex data. In general, AI aims to exhibit autonomous thinking and decision-making capabilities, while Cognitive Computing focuses on emulating and supporting human thinking and decision-making processes, and leverages this knowledge to design intelligent systems and technologies such as machine learning, Natural language processing (NLP), cognitive modeling and Brain-Computer Interfaces (BCIs). 

Advances in machine learning algorithms, particularly deep learning neural networks\cite{LeCun2015}, have revolutionized cognitive computing. These algorithms excel at recognizing patterns\cite{Shah2020}, making predictions\cite{Chimmula2020}, and learning from vast amounts of data. Deep learning has led to breakthroughs in image recognition\cite{Reddy20}, speech processing, and natural language understanding. NLP techniques have improved significantly, allowing machines to understand and generate human language more effectively\cite{Khurana:2023aa}. Cognitive informatics has played a crucial role in advancing NLP, enabling applications such as chatbots\cite{HALEEM2022100089}, and machine translation systems\cite{wu2016googles}.

Cognitive computing systems are increasingly being employed to assist in decision-making processes. They analyze data from various sources, apply reasoning algorithms, and provide insights to support human decision-makers. Cognitive computing-based decision support systems find applications in healthcare\cite{Zhang22a}, finance\cite{Chimmula2021} and other domains. Progress is made in cognitive modeling that simulate human cognitive processes\cite{lz18-2}. These models provide insights into decision-making, problem-solving, and learning mechanisms\cite{lz18-6}, enabling the creation of more intelligent and human-like systems. Bio-inspired artificial neural networks (ANNs) have been successfully employed for pattern recognition and prediction of simulated EEG\cite{Zhang2019d}. Previous studies have also highlighted the significance of both the ANN architecture and the quality of training data in influencing the performance of ANN training\cite{Zhang2019a}.  

Electroencephalography (EEG)-based BCIs utilize EEG signals to establish a direct communication pathway between the brain and external devices. By decoding and interpreting neural activity patterns, BCIs enable individuals to control external devices or interact with computer systems using their thoughts. Cognitive informatics plays a significant role in developing advanced algorithms for signal processing, feature extraction\cite{Ali16}, and cognitive state recognition\cite{Zhang2019e, Zhang2020} within EEG-based BCIs. Time-frequency domain analyses such as short-time Fourier transform \cite{lz17-4} was employed to capture the temporal and spectral features in EEG signals. The Spike Frequency Modulation (SFM) theory explains the nature of neural signals and their transformation in the nervous systems of the brain for neuroinformatics\cite{Wang2019}.

\subsection{SNN for Cognitive Computing}

Spiking neural networks (SNNs) have emerged as a promising approach in the field of cognitive computing. Inspired by the functioning of biological neurons, SNNs aim to model the spiking behaviuor of neurons and the timing of neural impulses, enabling more biologically plausible simulations of cognitive processes. There are some key aspects of spiking neural networks in the context of cognitive computing: First, unlike traditional artificial neural networks, which typically use continuous activation values, SNNs employ spiking neuron models that communicate through discrete, time-based spikes \cite{Chen2022, Izhikevich2003}. These spikes represent the timing and intensity of neural activity, allowing for more precise encoding and transmission of information \cite{Gerstner2002}.
Second, SNNs capture the temporal dynamics of information processing in the brain \cite{Zhang23a, Zhang22b}. By considering the timing of spikes, SNNs can represent and process information in a time-dependent manner, which is crucial for tasks involving temporal patterns\cite{Zhang22c} and timing-based computations. Third, the parallel architecture and discrete data representation of SNNs enable the integration of cognitive function simulation and hardware accleration. Modern programmable device like Field Programable Gates Arrays (FPGAs) have inherent parallel architectures and can provide hardware acceleration for SNNs, enabling real-time, low-power implementations of cognitive computing tasks \cite{JinZhang23}.

By leveraging the spiking behaviuor of neurons, temporal dynamics, and spike-based learning, spiking neural networks offer a promising avenue for cognitive computing. Their ability to model the intricate functions of the brain provides a more biologically plausible framework for understanding and emulating cognitive processes, opening doors for advancements in artificial intelligence and brain-inspired computing.

\section{Spiking Neural Network}

SNN is a type of neural network that models the behaviour of biological neurons more closely than traditional artificial neural networks. In SNNs, information is encoded in the form of spikes, which are discrete events that occur when the membrane potential of a neuron exceeds a certain threshold. The precise timing of the spikes can also carry information, as the relative timing of spikes between neurons can be used to represent the temporal sequence of events. This is in contrast to traditional ANN, where information is represented by continuous values. The behaviour of a spiking neuron is often modelled using a simple leaky integrate-and-fire model\cite{Teeter:2018aa}, which describes how the membrane potential of a neuron changes over time in response to inputs. When the membrane potential of a neuron reaches a certain threshold, a spike is generated and transmitted to other neurons in the network. SNNs are useful for modeling and simulating complex systems, such as the brain. They can also be used for tasks such as image classification of the numbers 0 to 9 \cite{KULKARNI2018,KASABOV2014}. However, training SNNs can be more difficult than traditional ANN, and there is still much research being done to improve the performance of these networks.

\subsection{SNN Encoding}

The encoding process typically involves transforming input signals into a series of spike trains that are fed into the network. This can be done using various encoding schemes, such as rate (frequency) encoding and temporal (phase) encoding. The choice of encoding scheme depends on the specific application and the nature of the input signals being processed.

\subsubsection{Rate encoding}
In rate encoding, the firing rate of a neuron is used to encode the strength or intensity of the input signal. For example, if a neuron has a high firing rate, it can be interpreted as receiving a strong input signal, while a low firing rate can be interpreted as a weak signal. To implement rate encoding in an SNN, the input signal is first transformed into a continuous variable and then mapped to a corresponding firing rate using a transfer function. This firing rate is then used to generate a series of spikes that represent the input signal. Rate encoding is simple to implement and requires less computational resources than temporal coding. It is relatively robust to noise and can work well with inputs that have a large dynamic range. However, it can not capture information about the precise timing of events in the input signals.  
 

\subsubsection{Temporal encoding}
In temporal coding, the timing of spikes is used to encode information. This can be done in various ways, such as using the precise timing of a single spike or using the relative timing of spikes between different neurons. For example, the input signal can be mapped to a sequence of spike times, where each spike time represents a specific feature or event in the signal. The timing of these spikes can then be used to represent the temporal sequence of events in the input signal.

Temporal encoding can also be implemented using population coding, where the input signal is represented by the activity of a population of neurons that fire at different times. In this case, the precise timing of individual spikes is less important, as the information is encoded in the overall pattern of activity across the population of neurons.

Temporal encoding can capture the precise timing of events in the input signal, which can be useful for tasks that require accurate timing, such as speech recognition or motor control. It is more efficient than rate encoding for certain types of inputs, such as those with sparse and rapidly changing dynamics. It can be used to represent complex temporal patterns, such as sequences or rhythms. However, it is more computationally expensive than rate encoding, as it requires more precise spike timing information.

%
%

\subsection{Training and Decoding}
Once the information is encoded as spikes, it is transmitted through the network using synaptic connections between neurons. The strength of these connections can be adjusted through a training process, which allows the network to adapt and learn. 
The adaptability of SNN is analogous to the synaptic plasticity observed in biological neural networks. 

The decoding process involves analyzing the spiking activity of the neurons in the network to extract useful information. This process involves reading out the firing rate or the precise timing of spikes in certain neurons or groups of neurons. The decoded information can then be used to make predictions, control systems, or perform other tasks depending on the specific application.



\subsection{Spiking Neuron based on Complex Exponential Function and SNN based on DFT}

The proposed SNN is inspired by the well-established practice of constructing arbitrary time series signals using Fourier Series. In equation (\ref{eq:SNN}), a spiking neuron $s(t)$ is expressed as a complex exponential function; and an SNN is constructed by N spiking neurons. Each spiking neuron is assigned an integer multiple of the fundamental frequency $f_0$: $f_k = k f_0$, representing the frequency components in the DFT. The DFT coefficient $C_k$ equals the multiplication of the amplitude A and the initial phase $\phi$ of the k-th spiking neuron. The equation of the SNN resembles the inverse DFT equation used for reconstructing time signals from DFT coefficients. 

\begin{equation}
\label{eq:SNN}
\begin{aligned}
s(t) &= A e^{i(2\pi f t + \phi)} = C e^{i2\pi f t}, C = A e^{i\phi} \\
SNN &= \sum_{k = 0}^{N-1} s_k[t]= \sum_{k = 0}^{N-1} C_k e^{i 2\pi f_k t} 
\end{aligned}
\end{equation}

\subsection{Derivative of the Frequency Spectrum and Phase Difference}

The derivative of the frequency spectrum with respect to frequency describes how the amplitude or intensity of different frequency components in a signal changes as the frequency changes. The differential equation representing this relationship is shown in (\ref{eq:DiffDFT}).

\begin{equation}
\label{eq:DiffDFT}
\begin{aligned}
Y[k] &= \sum^{N-1}_{n=0} x[n] e^{-i \frac{2\pi}{N}kn} \\
\Rightarrow \frac{Y[k]}{dk} &= \sum^{N-1}_{n=0} x[n] \frac{d}{dk}\left(e^{-i \frac{2\pi}{N}kn}\right) \\ 
&= \sum^{N-1}_{n=0} x[n] \left(-i\frac{2\pi}{N}n\right) e^{-i \frac{2\pi}{N}kn}
\end{aligned}
\end{equation}

$Y[k]$ is the DFT coefficient corresponding to the frequency bin $k$, $x[n]$ is the discrete time domain signal and N is the number of samples. This equation indicates that the derivative of the frequency spectrum with respect to frequency is equal to the derivative of the DFT coefficients with respect to the frequency bin index k. In other words, it relates how changes in the frequency index affect the spectrum of the signal. The expression $\Delta \phi= -i\frac{2\pi}{N}n$ represents the phase difference between adjacent frequency coefficients. Hence the derivative of the frequency spectrum with respect to frequency is proportional to the phase difference $\Delta \phi$ between adjacent frequency coefficients. 

\subsection{Gradient of Frequency Spectrum in Polar Coordinate}

The derivative describes the rate of change of a function with respect to a single variable, while the gradient represents the collection of partial derivatives of a multivariable function, indicating the direction and steepness of change in a multidimensional space. The gradient of the frequency spectrum can be represented in the form of partial derivative in (\ref{eq:PolarPartialDiff}) with respect to the magnitude $r$ and phase $\theta$. 

\begin{equation}
\label{eq:PolarPartialDiff}
\begin{aligned}
dY(r,\theta) &=\frac{ \partial Y}{\partial r} d r + \frac{ \partial Y}{\partial \theta} d \theta \\
dY &= (\triangledown Y)_r ds_r +  (\triangledown Y)_\theta ds_\theta =  (\triangledown Y) \cdot ds\\
 \triangledown Y & =\frac{ \partial Y}{\partial r} u_r + \frac{ \partial Y}{r\partial \theta}  u_\theta 
\end{aligned}
\end{equation}

$(\triangledown Y)_r$ is the partial derivative of Y with respect to r, while $(\triangledown Y)_\theta$ is the partial derivative of Y with respect to $\theta$. The distance in polar coordinates upon making small changes in the variables r and $\theta$ is denoted as $ds^2 = dr^2 +r^2 d\theta^2, ds_r = dr, ds_\theta = rd\theta$. Because $ds_\theta $ has a factor of r in it, there is a compensating factor of r in the denominator of the component of $(\triangledown Y)$ in the $\theta$ direction: $ (\triangledown Y)_\theta = \frac{\partial Y}{r \partial \theta}$. The gradient of $\theta$ is $\frac{u_\theta}{r}$, the gradient of $\frac{1}{r}$ is $-\frac{u_r}{r^2}$.

\section{Phase Encoding for Spiking Train}

The DFT-based SNN gives rise to a novel phase encoding scheme. The delay of an impulse signal is proportional to the phase delay of the frequency coefficients. The time shifting property of the Fourier transform listed in (\ref{eq:FTTimeShifting}) can be used to mathematically represent the spike delay encoding.

\begin{equation}
\label{eq:FTTimeShifting}
x[n-n_0] \Leftrightarrow e^{-i\frac{2\pi}{N}k n_0} X[k]
\end{equation}

The delayed impulse function is used to illustrate the SNN phase encoding scheme. The DFT of a delayed impulse function is expressed as (\ref{eq:DFTDeltaDelay}). The DFT coefficients Y[k] are used to specify the complex exponential spiking neurons $s_k$ in the SNN. 

\begin{equation}
\label{eq:DFTDeltaDelay}
\begin{aligned}
Y[k]&=\boldsymbol{DFT}\{\delta[n-n_0]\} = \sum_{n=0}^{N-1} \delta[n-n_0] e^{-i \frac{2\pi}{N} nk} \\
&= e^{-i \frac{2\pi}{N} kn_0} , \quad k = 0,1,2,...,N-1
\end{aligned}
\end{equation}

Assume the magnitude of the spiking neurons are 1. The natural logarithm of the ratio between two adjacent frequency coefficients equals the phase difference (\ref{eq:PhaseDiff}).

\begin{equation}
\label{eq:PhaseDiff}
\begin{aligned}
\ln \frac{C[k+1]}{C[k]} & = \ln  e^{-i\frac{2\pi}{N} n_0} = \ln  e^{ \Delta\phi} = \Delta\phi \\ \Delta\phi & = -i\frac{2\pi}{N} n_0
\end{aligned}
\end{equation}

\subsection{Phase Encoding of Delayed Impulse (N=2)}

\begin{figure}[thb]
\centering
\includegraphics[width=1\linewidth]{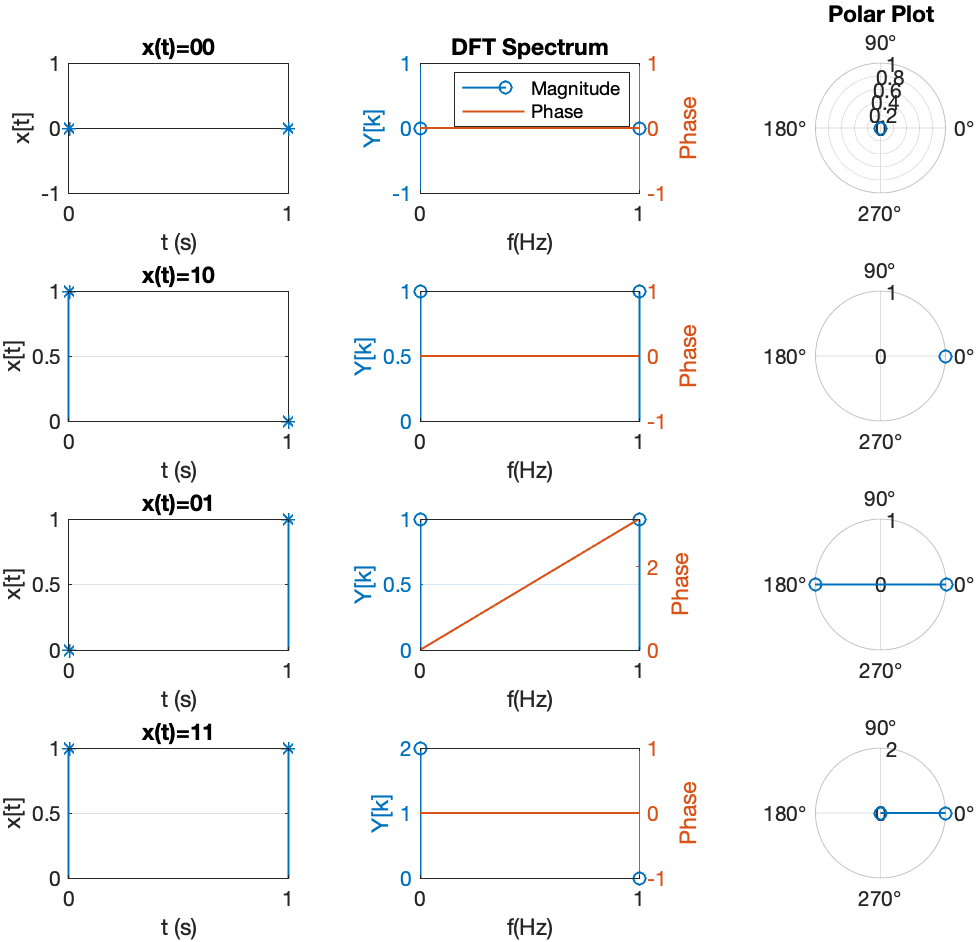}
\caption{Impulse Patterns (N=2), DFT spectrum and Polar Plot }
\label{fig:2NImpulseDFTPolar4x3}
\end{figure}

For a discrete signals x(t) with two samples (N=2), there are four possible patterns: 00,10,01,11. Fig. \ref{fig:2NImpulseDFTPolar4x3} shows the four sequential patterns of the signal x(t). Since the length of the impulse signal is 2, an N=2 DFT is applied to the four time sequences. The DFT spectrum, including both magnitude and phase, is plotted in the second column. The two DFT coefficients define the oscillating frequencies and initial state of two spiking neurons in an SNN. The polar plots of the DFT spectrum are shown in the third column, indicating the magnitude and phase of each coefficient.

Table. \ref{tab:N2Encoding} provides a numerical explanation of the relationship between the impulse delays $n_0$ and the phase pattern of the SNN. The magnitude of both spiking neurons is 1. The spiking neuron $s_k$ represents the frequency component at $\omega_k = k\Omega_0$. For the two impulse signals: ``10" and ``01", the phase of the two spiking neurons increases linearly with an increment defined by the delay: $\Delta \phi =-i \frac{2\pi}{N} n_0 =-i \Omega_0 n_0$. Therefore the impulse delay can be encoded as $n_0 = i\frac{\Delta\phi}{\Omega_0}$.

\begin{table}[h]
\caption{SNN Phase Increment $\Delta \phi$ Encoding for Delayed Impulse: $N=2, \Omega_0 =\pi, \omega =\Omega_0 k, k = 0,1$}
\centering
\begin{tabular}{|c|c|c|c||c|}
\hline
\bf x(t) & $n_0$ & Y0 & Y1  &  $\Delta \phi$ \\
\hline
00 & 0&  0 & 0 & 0 \\
\hline
\bf 10 & 0 & 1 & 1 & 0 \\
\hline
\bf 01 & 1& 1  & $1\angle -\pi$ & $-\pi$ \\
\hline
11 & 0 & 2 & 0 & 0 \\
\hline
\end{tabular}
\label{tab:N2Encoding}
\end{table}

Since s0 is a constant, when N=2, the SNN only consists of one non-zero frequency component $s1= e^{i\pi t}$, with an angular frequency of $\pi$ radians/sec. Hence, the impulse signal x(t)=``10" with a delay of 0 is encoded as $\Delta \phi = 0$, or $2\pi$, and the the impulse signal ``x(t)=01" with a delay of 1s is encoded as $\Delta \phi = \pi$. It should be noted that delay encoding is not applicable to signal patterns consisting entirely of zeros or ones, as there is no delay information to encode. 

\begin{figure}[t]
\centering
   \begin{subfigure}{0.45\linewidth}
   \centering
   \includegraphics[width=\linewidth]{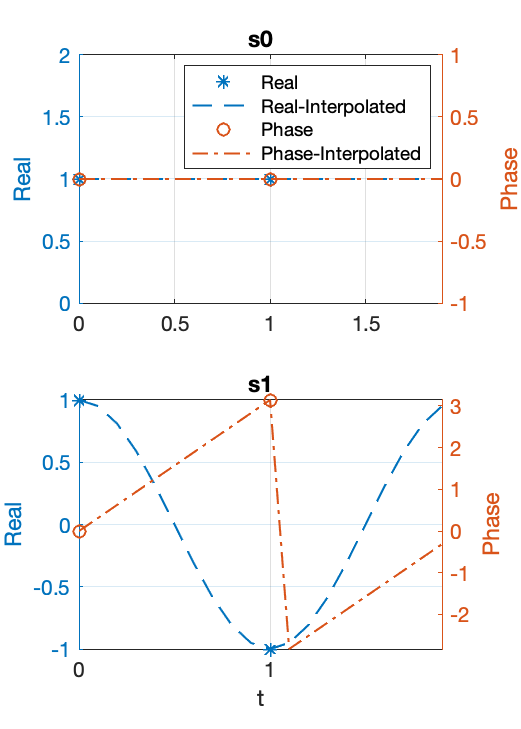}
   \caption{x(t)=``10"}
   \label{fig:2x1SNNImpulseDelay0}
\end{subfigure}
\hfill
\begin{subfigure}{0.45\linewidth}
   \centering
   \includegraphics[width=\linewidth]{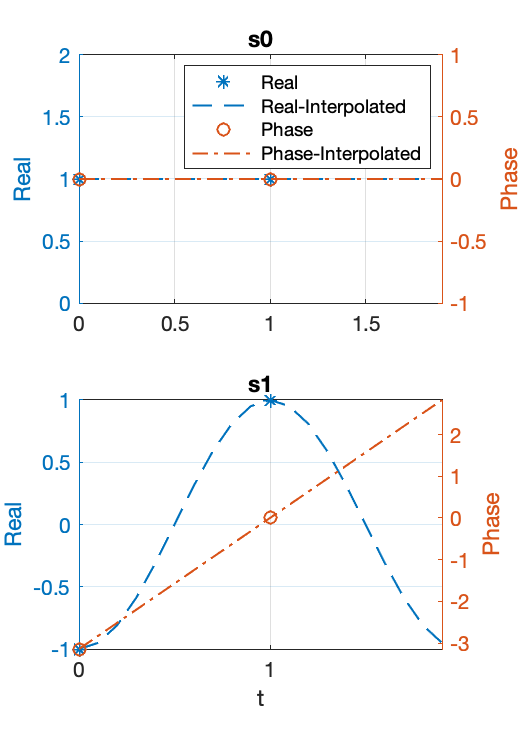}
   \caption{x(t)=``01"}
   \label{fig:2x1SNNImpulseDelay1}
\end{subfigure}
\centering
\caption{SNN (N=2) Reconstruction of Delayed Impulses}
\end{figure}

{\it Signal Reconstruction Using SNN:} The number of spiking neurons ($\omega \neq 0$) required to to represent an N-point DFT of an N time points signal is N-1. The neuron s0 with frequency f=0 represents a constant value in the SNN. When N=2, one additional spiking neuron is required with oscillating frequency $\omega = \pi$. The original impulse signal can be reconstructed by summing all the spiking neurons. Figures \ref{fig:2x1SNNImpulseDelay0} and \ref{fig:2x1SNNImpulseDelay1} show the spiking neurons show the spiking neurons s0 and s1 of the N2-SNN for the impulse signal x(t)=``10" with 0 delay, and and x(t)=``01" with 1s delay. The real component and phase of the spiking neurons are interpolated by a factor of 10. The initial phase of s1 is 0 for x(t)=``10", and $-\pi$ for x(t)=``01". The actual samples are indicated by marker `*' for the real components, and `o' for the phase. The sum of these two neurons reconstructs the original impulse signal. 

\subsection{Phase Encoding of Delayed Impulse (N=4)}

\subsubsection{Phase Encoding of Delayed Single Impulse}

\begin{figure*}[thb]
\centering
\begin{minipage}{.48\textwidth}
\centering
\includegraphics[width=1\linewidth]{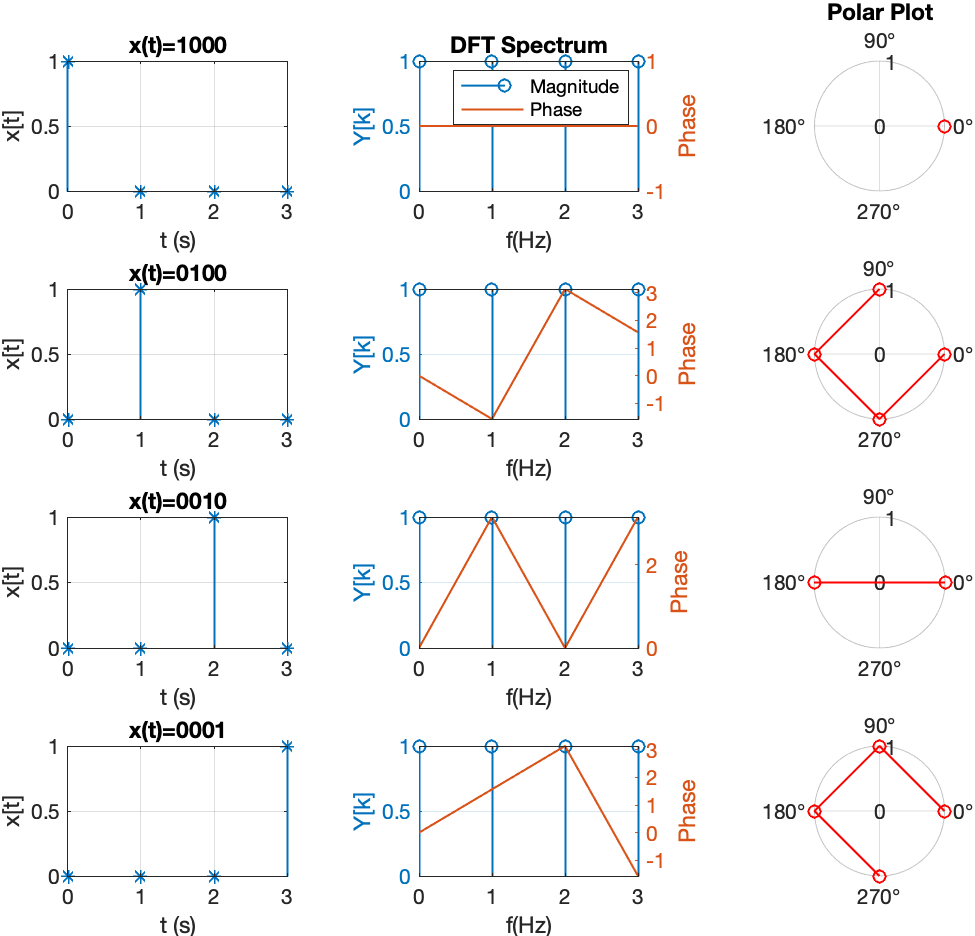}
\caption{Delayed Simple Impulse (N=4), DFT and Polar Plot}
\label{fig:4x3OneDelayedImpulse}
\end{minipage}
\quad
\begin{minipage}{0.48\textwidth}
\centering
\includegraphics[width=1\linewidth]{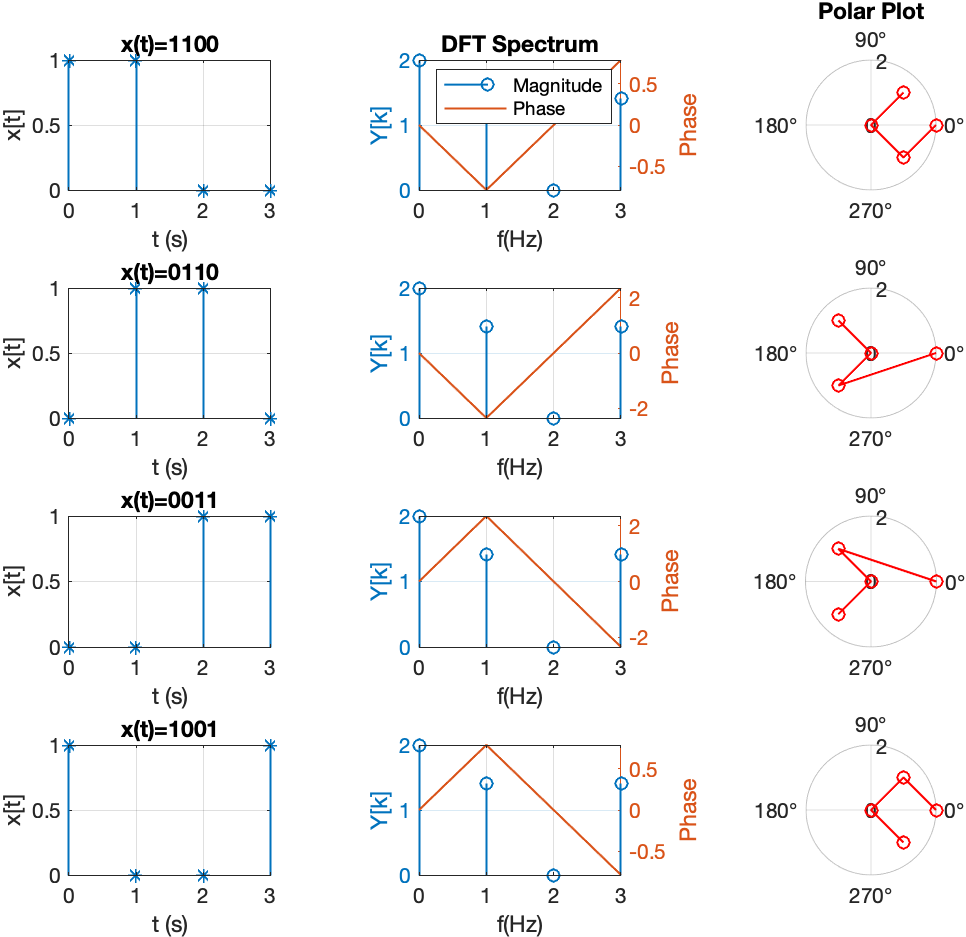}
\caption{Delayed Two Impulses(N=4), DFT and Polar Plot}
\label{fig:4x3TwoDelayedImpulse}
\end{minipage}
\end{figure*}

Fig. \ref{fig:4x3OneDelayedImpulse} illustrates a discrete impulse generated with period T=4s. The impulse signal is delayed by 1 to 3 seconds, respectively, and displayed in the first column. Considering the length of the impulse signal is 4, an N=4 DFT is applied. The DFT spectrum including both magnitude and phase is plotted in the second column. The 4 DFT coefficients plotted in the DFT spectrum represent 4 spiking neurons oscillating at 4 distinct frequencies. It can be observed that the magnitudes of the impulse train in the DFT spectrum are all 1. However, the phase spectrum varies depending on the delay of the impulse. The derivative of the DFT spectrum with respect to the frequency f is equivalent to the phase difference across the 4 spiking neurons. The polar plots, displayed in the third column, indicate the phase of each spiking neuron (frequency component). Although the magnitude of the four DFT spectrum coefficients for the four x(t) impulse signals are all 1, their phase patterns differ. The polar plot of the 4 spiking neurons exhibits a linear phase difference, which corresponds to the sequential firing pattern associated with the original delayed impulse. 

Table. \ref{tab:SNNPhaseIncrementDelayedImpulse} lists the initial phase $\angle Y_k$ and phase difference $\Delta \phi= -i\frac{2\pi}{N} n_0$, elucidating the relationship between the impulse delays $n_0$ and the phase difference $\Delta\phi$. The magnitude of the 4 spiking neurons are all 1. The spiking neuron $s_k$ represents the frequency component at $\omega_k = k\Omega_0$. The phase of the 4 spiking neurons increases linearly with an increment defined by the delay $n_0$ and the fundamental oscillating frequency $\Omega_0$ : $\Delta \phi = -in_0\Omega_0  = -in_0\pi /2$.

\begin{table}[h]
\caption{SNN Phase Increment $\Delta \phi$ Encoding for Delayed Single Impulse: $N=4, \Omega_0 =\pi/2, \omega =\Omega_0 k, k = 0,1,2,3$}
\centering
\begin{tabular}{|c|c|l|l|l|l||c|}
\hline
\bf x(t) & $n_0$ & Y0 & Y1  & Y2 & Y3 &  $\Delta \phi$ \\
\hline
1000 & 0 & $1\angle 0$ & $1\angle 0$ & $1\angle 0$ &$1\angle 0$ & 0 \\
\hline
0100 & 1 & $1\angle 0$ & $1\angle \frac{-\pi}{2}$  & $1\angle -\pi$  &  $1\angle  \frac{-3\pi}{2}$ & $ \frac{-\pi}{2}$ \\
\hline
0010 & 2 & $1\angle 0$ & $1\angle -\pi$  & $1\angle -2\pi$  &  $1\angle -3\pi$ & $-\pi$ \\ 
\hline
0001 & 3 & $1\angle 0$ & $1\angle \frac{-3\pi}{2}$  & $1\angle \frac{-6\pi}{2}$  &  $1\angle \frac{-9\pi}{2}$ & $-3\pi/2$ \\
\hline
\end{tabular}
\label{tab:SNNPhaseIncrementDelayedImpulse}
\end{table}

Note that the phase of a complex exponential function is periodic: $e^{i\theta} = e^{i(\theta +2\pi k)}$, where k is an integer. The range of the $2\pi$ period are set as $[0,2\pi)$ or $(-\pi,\pi]$ by default in MATLAB. In the DFT spectrum plot, the phase range is set to $(-\pi,\pi]$. In the polar plot, the range of the phase is set to $[0,2\pi)$. Therefore, the $\Delta \phi = -\pi$ for the delayed impulse x(t)= ``0010" is equivalent to $\pi$. Similarly, the $\Delta \phi = \frac{-3\pi}{2}$ for the delayed impulse x(t)= ``0001" is equivalent to $\frac{\pi}{2}$. The phases of the DFT coefficients are adjusted to be within the default range. 

\begin{figure*}[!h]
\centering
\begin{subfigure}{0.24\linewidth}
   \centering
   \includegraphics[width=\linewidth]{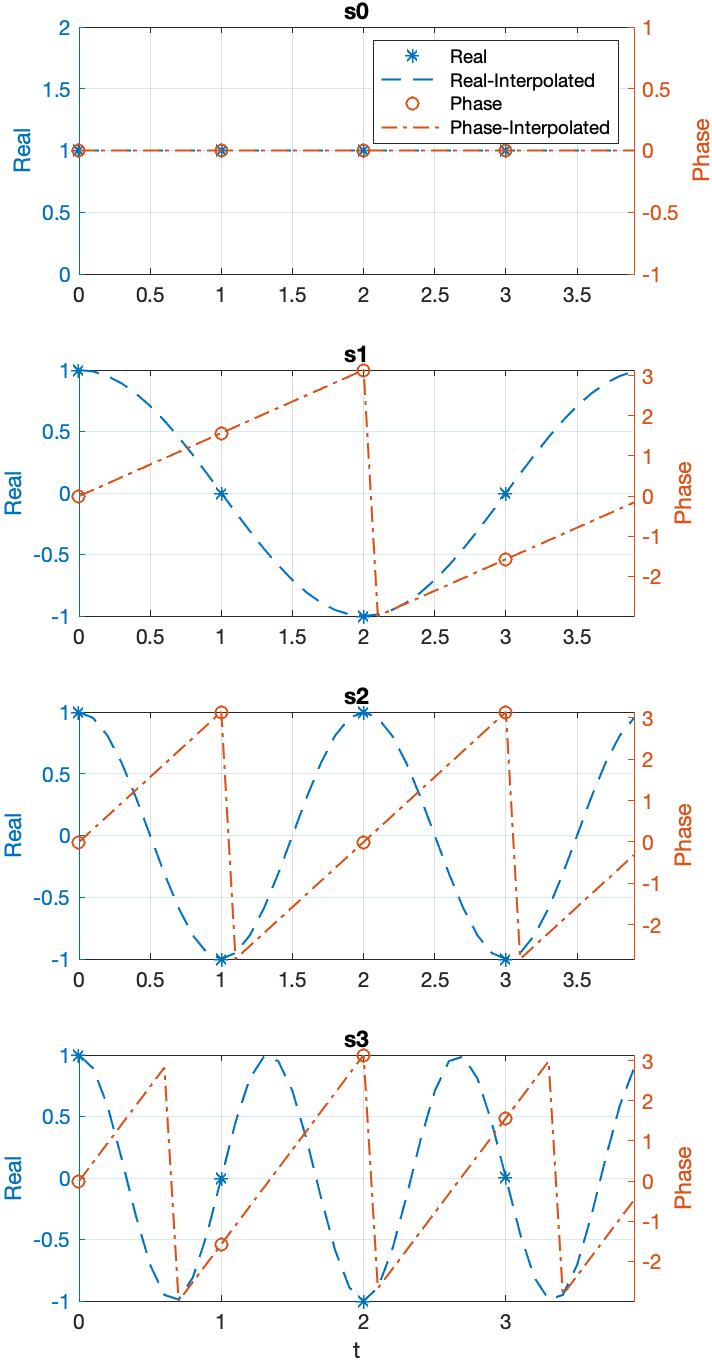}
   \caption{x(t)=``1000"}
   \label{fig:4x1SNNImpulseDelay0}
\end{subfigure}
\hfill
\begin{subfigure}{0.24\linewidth}
   \centering
   \includegraphics[width=\linewidth]{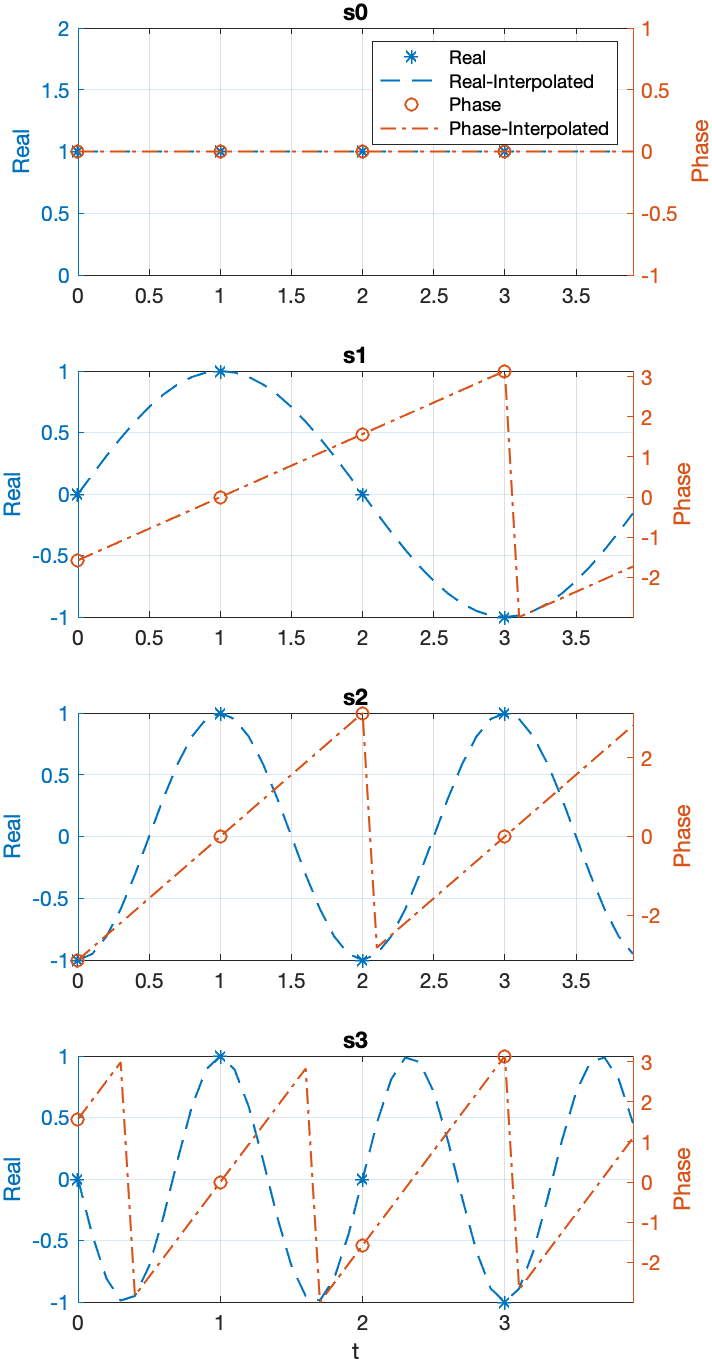}
   \caption{x(t)=``0100"}
   \label{fig:4x1SNNImpulseDelay1}
\end{subfigure}
\hfill
\begin{subfigure}{0.24\linewidth}
   \centering
   \includegraphics[width=\linewidth]{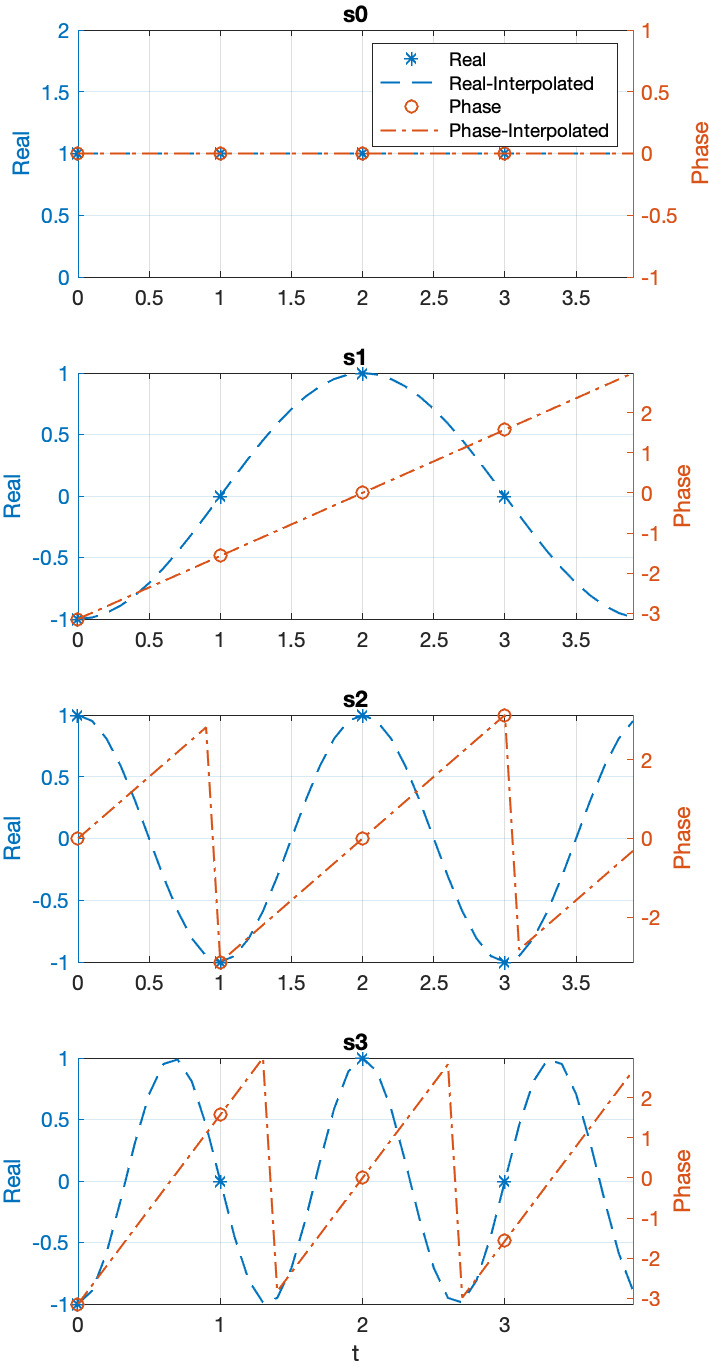}
   \caption{x(t)=``0010"}
   \label{fig:4x1SNNImpulseDelay2}
\end{subfigure}
\hfill
\begin{subfigure}{0.24\linewidth}
   \centering
   \includegraphics[width=\linewidth]{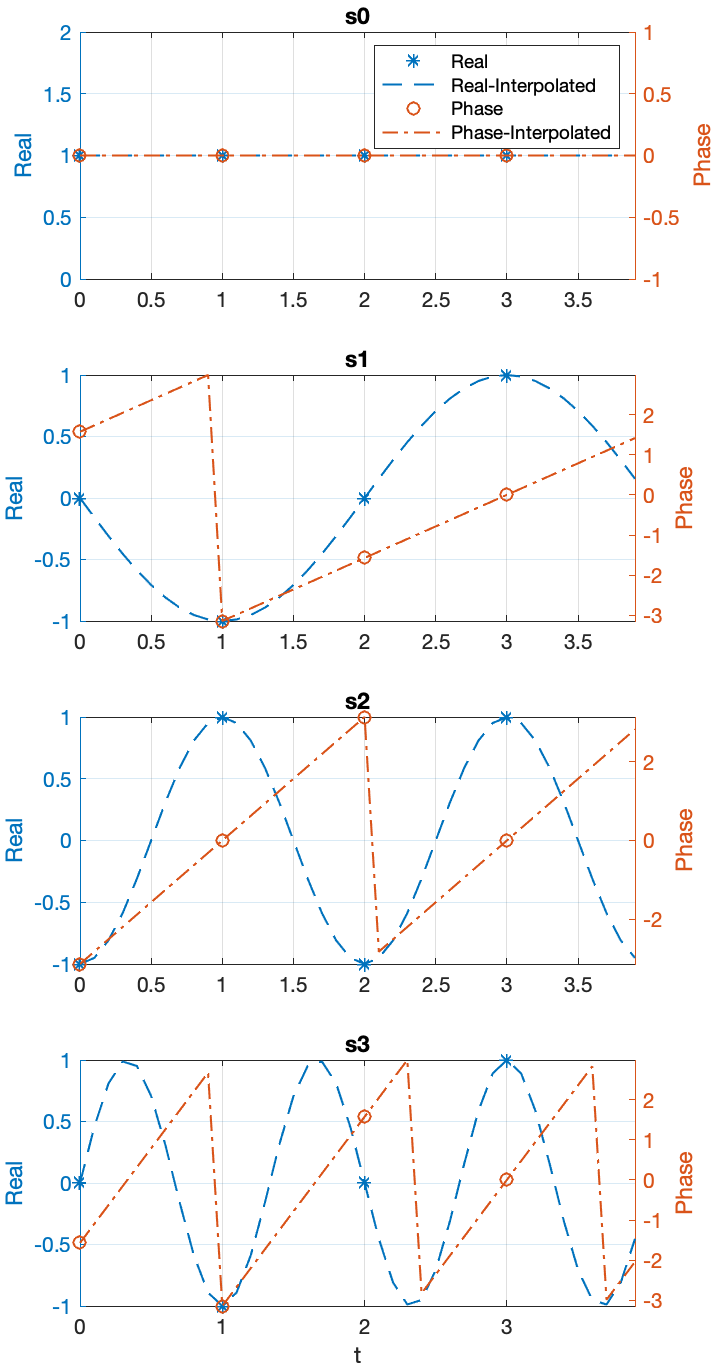}
   \caption{x(t)=``0001"}
   \label{fig:4x1SNNImpulseDelay3}
\end{subfigure}
\centering
\caption{SNN (N=4) Representation of Single Delayed Impulse}
\end{figure*}

{\it Signal Reconstruction Using SNN:} The original impulse signal can be reconstructed by summing four spiking neurons. Fig. \ref{fig:4x1SNNImpulseDelay0} to \ref{fig:4x1SNNImpulseDelay3} illustrate four spiking neurons used to reconstruct an impulse signal with time delay of $n_0 =0,1,2,$ and $3$, respectively. The plots for the real components and phases of the four spiking neurons are interpolated by a factor of 10. The actual four samples within one period are denoted by the marker `*' for the real components, and `o' for the phase. The sum of these four neurons reconstructs the original impulse signals. The neuron s0 indicates a constant value with a frequency of 0. The initial phase of s1, s2 and s3 are determined by the DFT coefficients Y1, Y2 and Y3 respectively, as listed in Table.\ref{tab:SNNPhaseIncrementDelayedImpulse}, and depicted in the DFT spectrum and the polar plots in Fig.\ref{fig:4x3OneDelayedImpulse}. 

\begin{figure*}[!h]
\centering
\begin{subfigure}{0.24\linewidth}
   \centering
   \includegraphics[width=\linewidth]{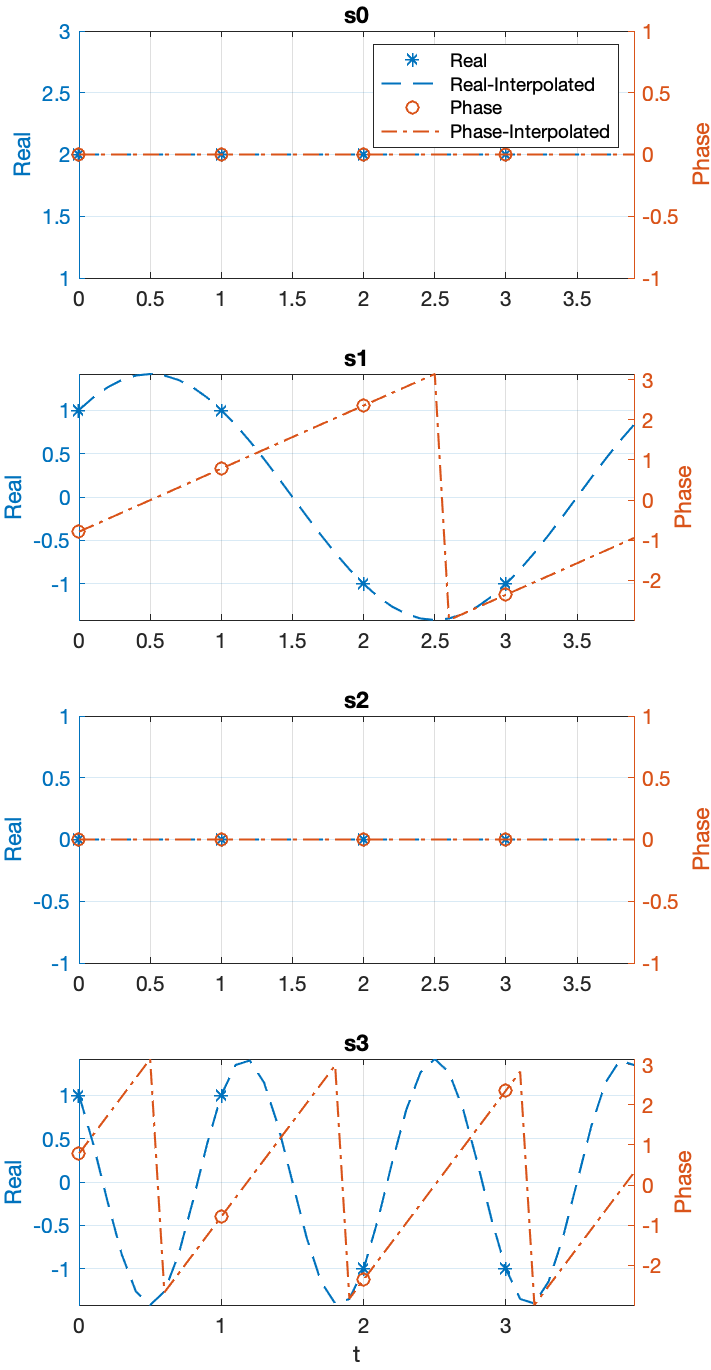}
   \caption{x(t)=``1100"}
   \label{fig:4x1SNN2spikesDelay0}
\end{subfigure}
\hfill
\begin{subfigure}{0.24\linewidth}
   \centering
   \includegraphics[width=\linewidth]{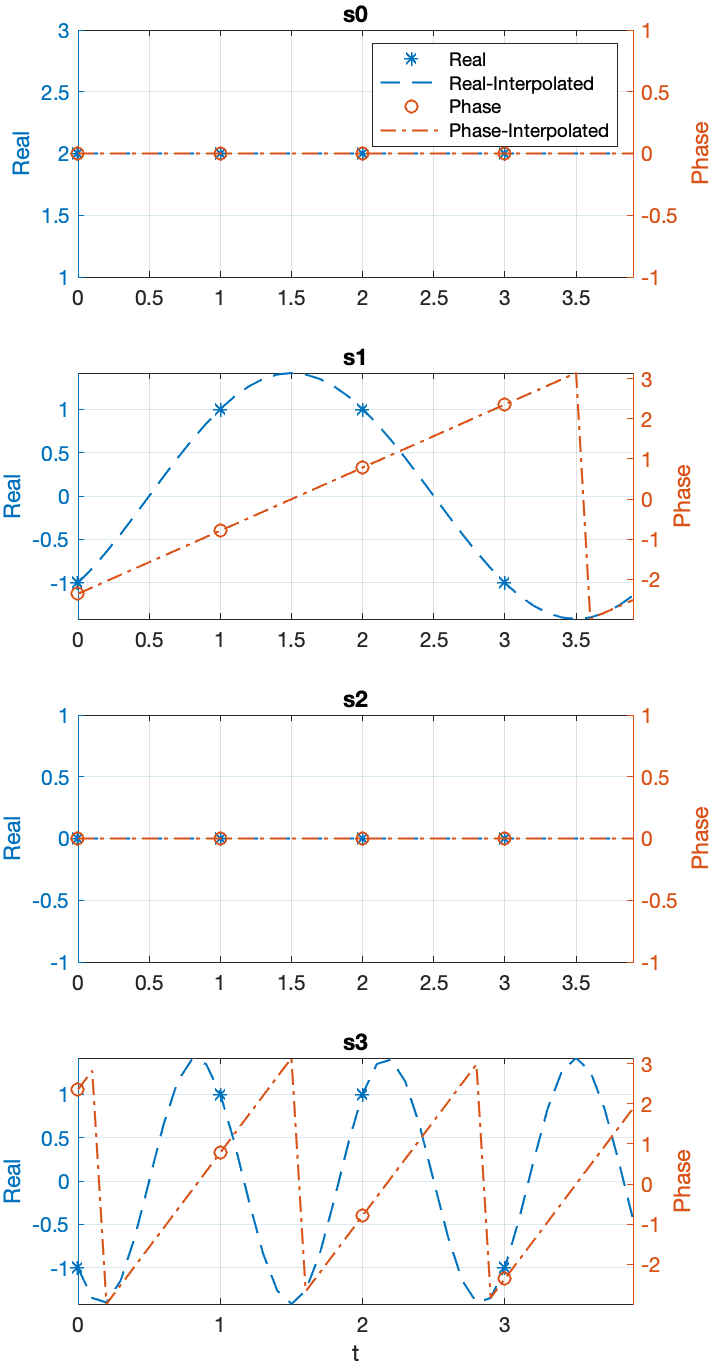}
   \caption{x(t)=``0110"}
   \label{fig:4x1SNN2spikesDelay1}
\end{subfigure}
\hfill
\begin{subfigure}{0.24\linewidth}
   \centering
   \includegraphics[width=\linewidth]{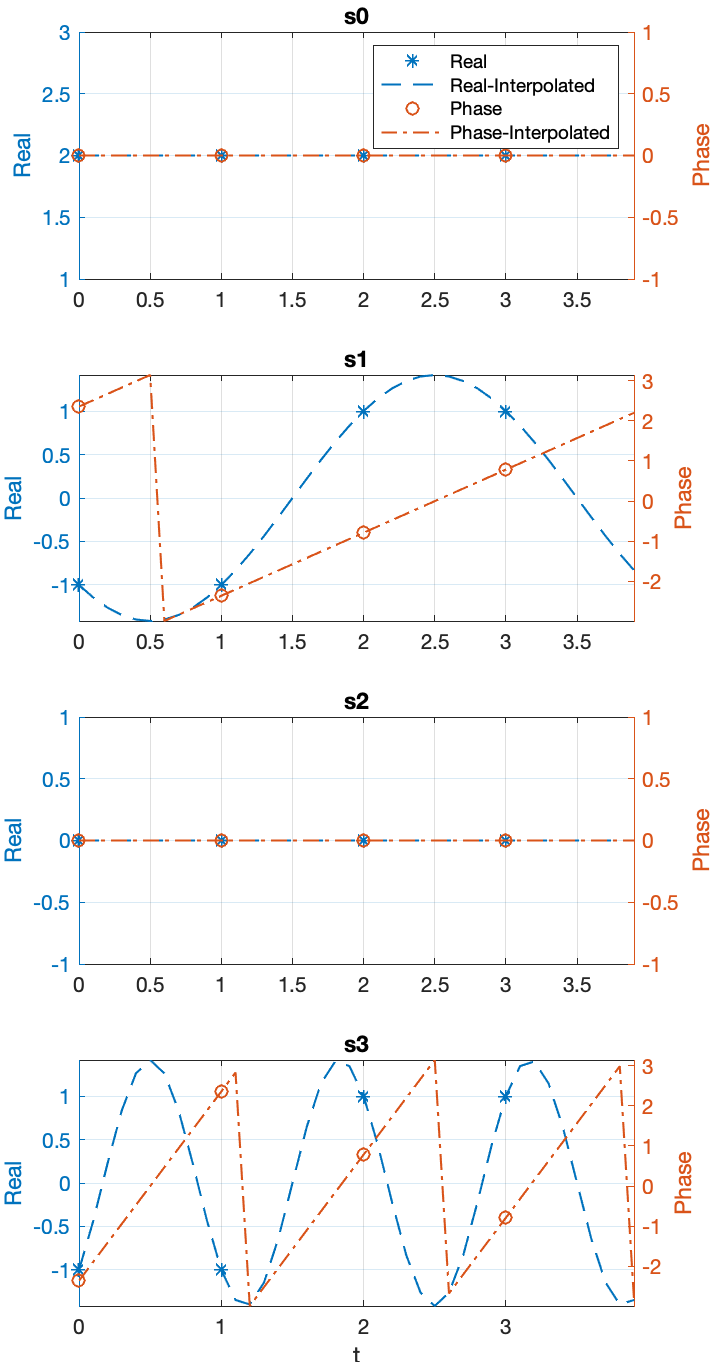}
   \caption{x(t)=``0011"}
   \label{fig:4x1SNN2spikesDelay2}
\end{subfigure}
\hfill
\begin{subfigure}{0.24\linewidth}
   \centering
   \includegraphics[width=\linewidth]{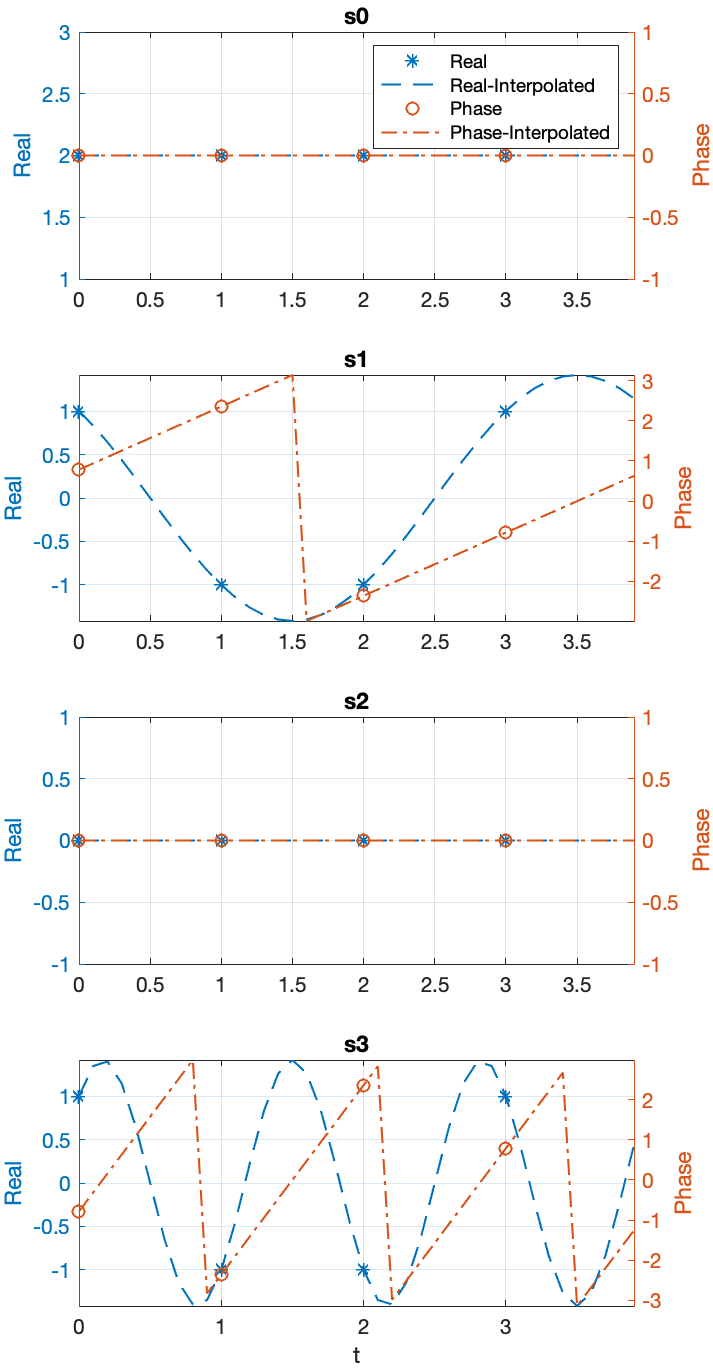}
   \caption{x(t)=``1001"}
   \label{fig:4x1SNN2spikesDelay3}
\end{subfigure}
\centering
\caption{SNN (N=4) Representation of Two Delayed Impulse}
\end{figure*}

\begin{figure*}[h]
\centering
\begin{minipage}{.48\textwidth}
\centering
\includegraphics[width=1\linewidth]{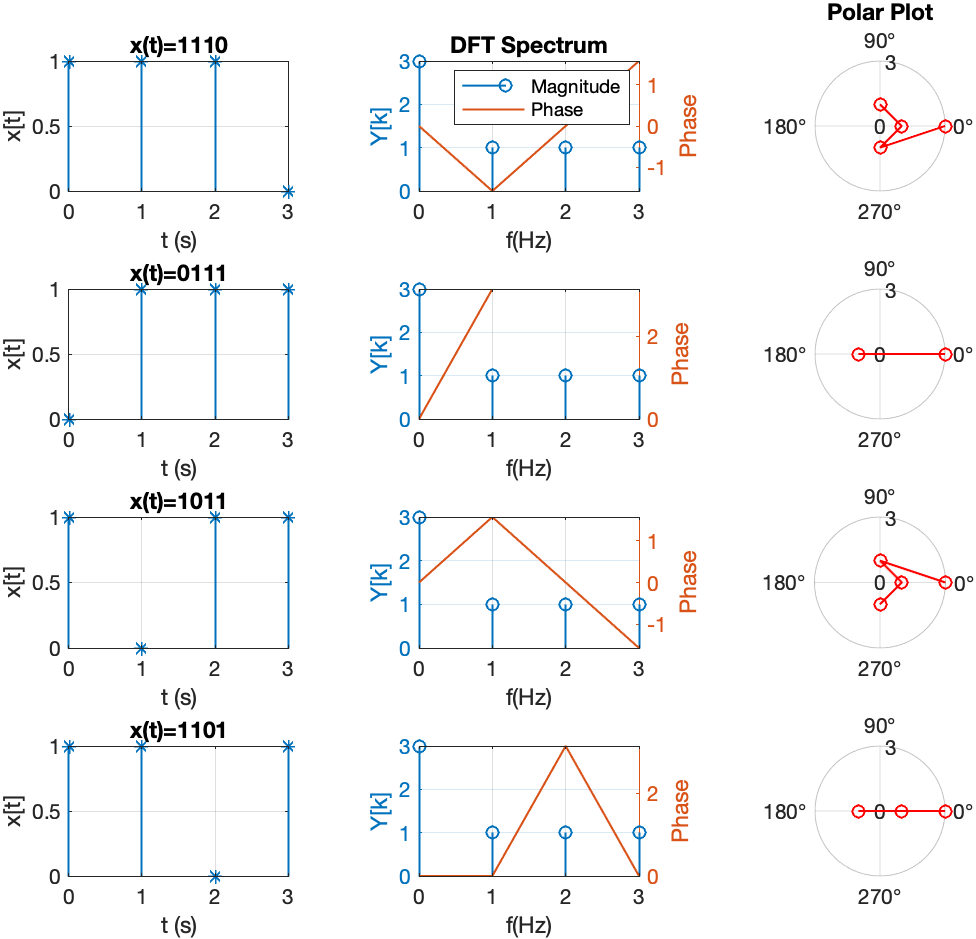}
\caption{Delayed 3 Impulses (N=4), DFT and Polar Plot}
\label{fig:4x3ThreeDelayedImpulse}
\end{minipage}
\quad
\begin{minipage}{0.48\textwidth}
\centering
\includegraphics[width=1\linewidth]{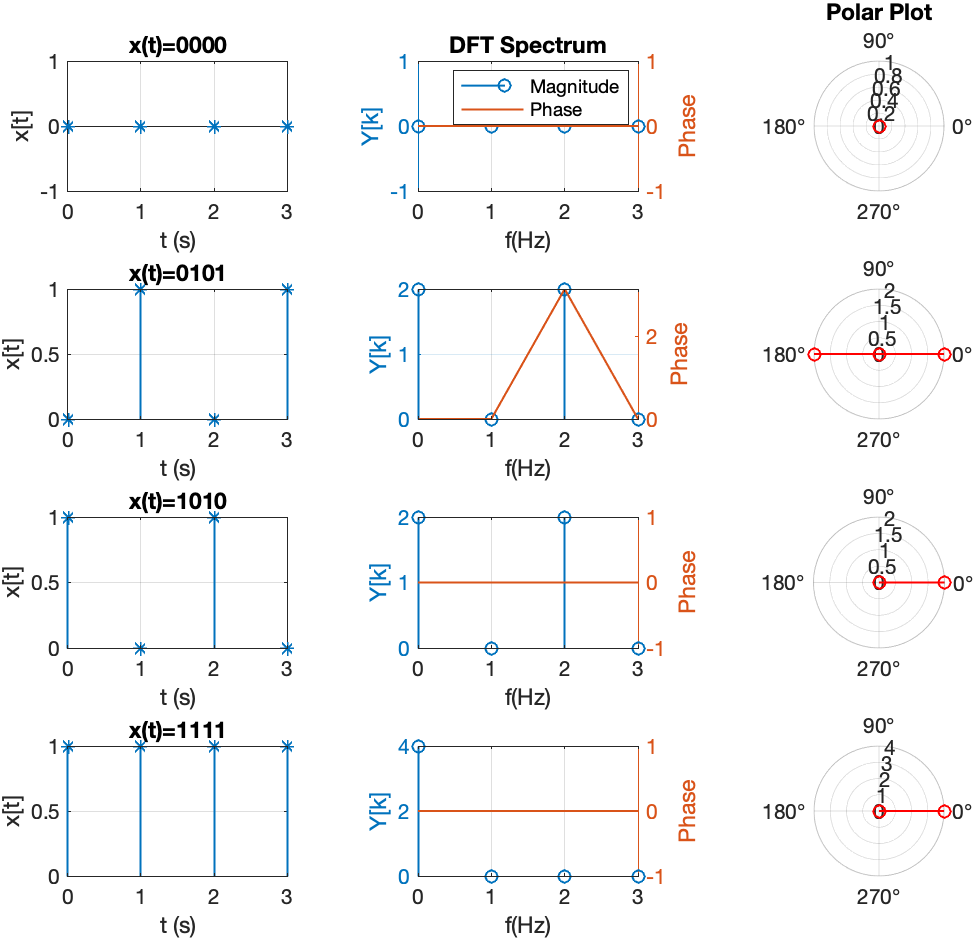}
\caption{Other Impulses Patterns(N=4), DFT and Polar Plot }
\label{fig:4x3FourLeftPatterns}
\end{minipage}
\end{figure*}

\begin{figure*}[!h]
\centering
\begin{subfigure}{0.24\linewidth}
   \centering
   \includegraphics[width=\linewidth]{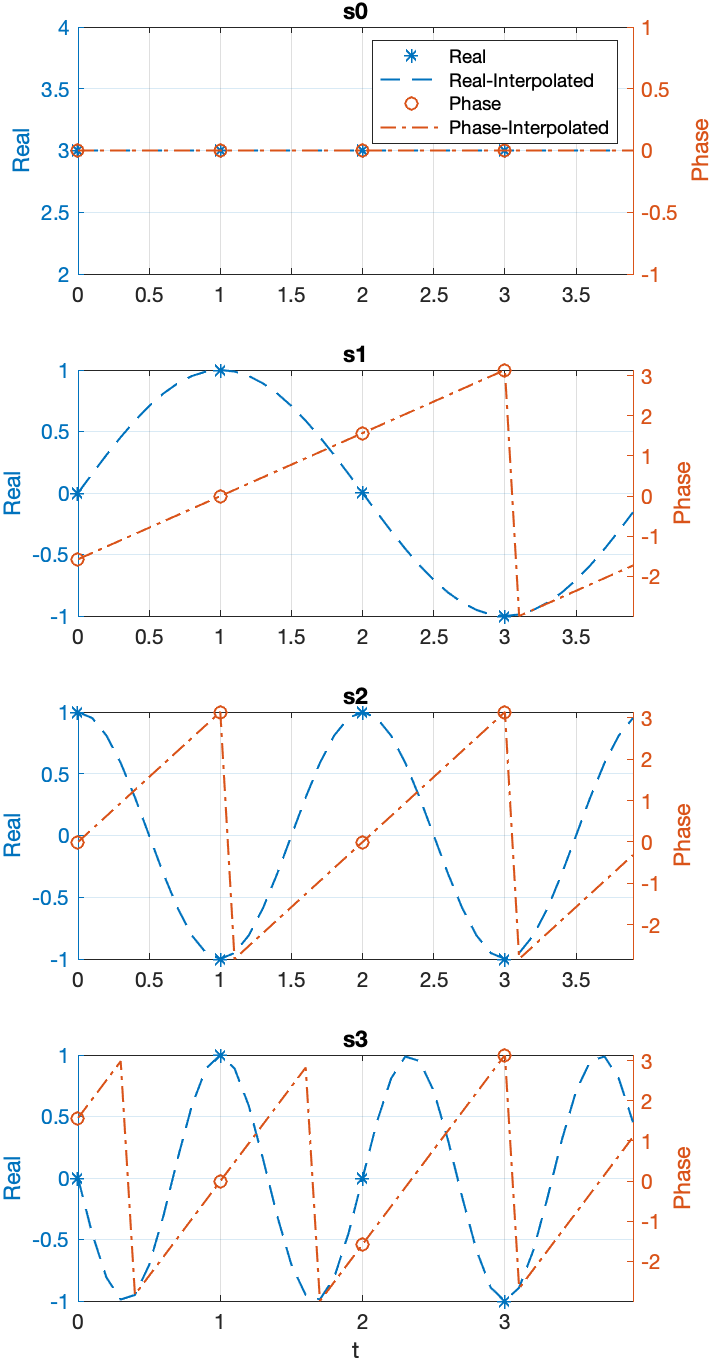}
   \caption{x(t)=``1110"}
   \label{fig:4x1SNN3spikesDelay0}
\end{subfigure}
\hfill
\begin{subfigure}{0.24\linewidth}
   \centering
   \includegraphics[width=\linewidth]{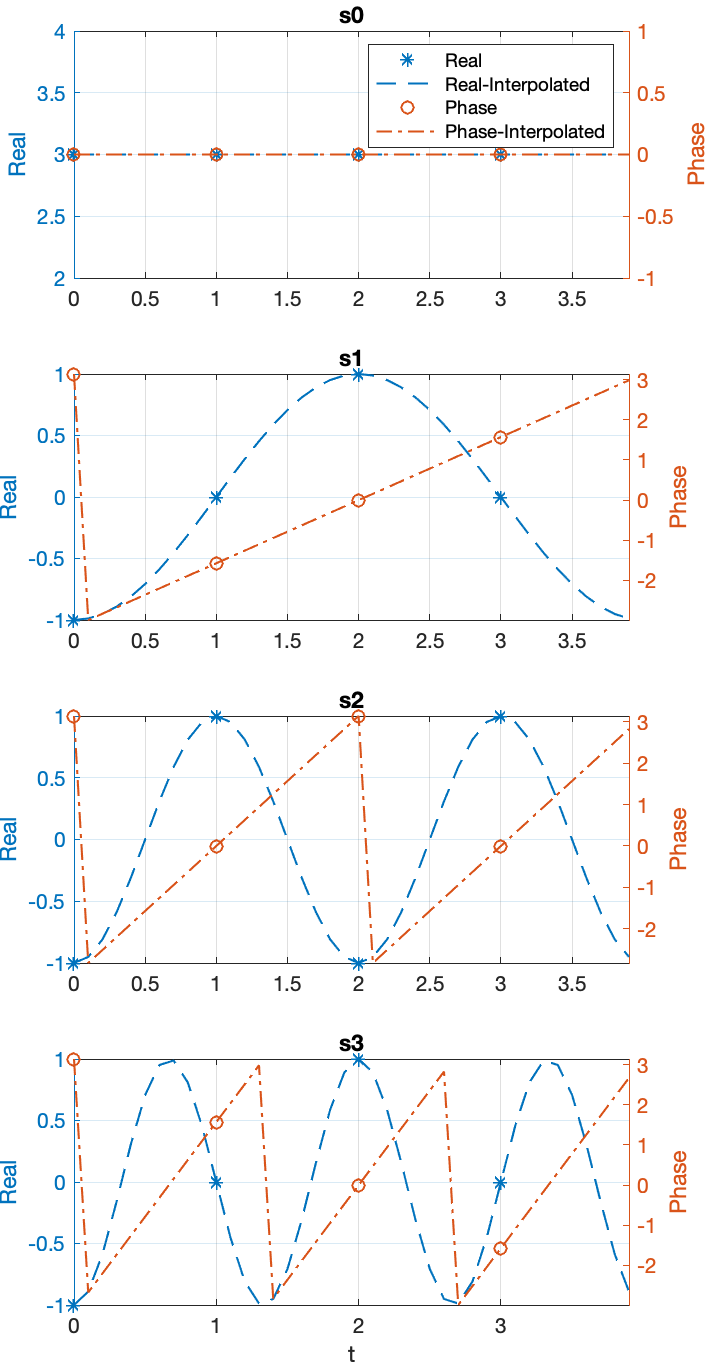}
   \caption{x(t)=``0111"}
   \label{fig:4x1SNN3spikesDelay1}
\end{subfigure}
\hfill
\begin{subfigure}{0.24\linewidth}
   \centering
   \includegraphics[width=\linewidth]{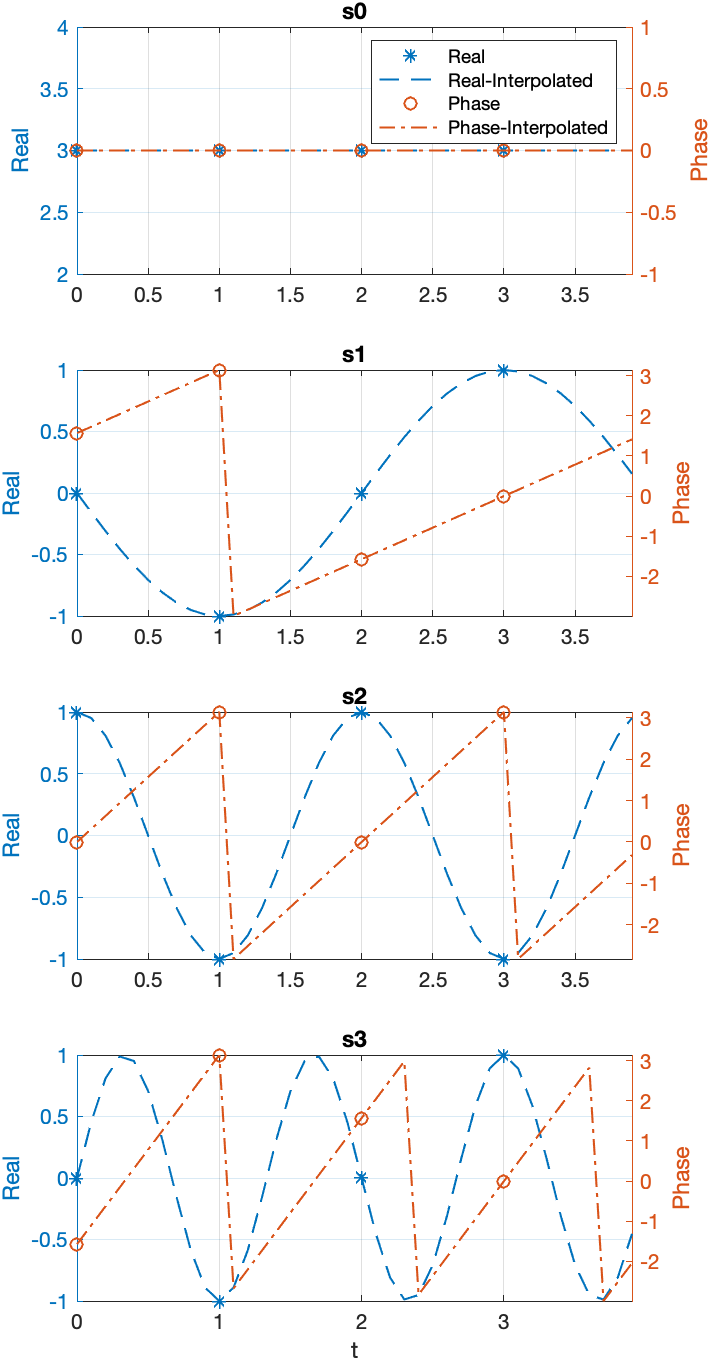}
   \caption{x(t)=``1011"}
   \label{fig:4x1SNN3spikesDelay2}
\end{subfigure}
\hfill
\begin{subfigure}{0.24\linewidth}
   \centering
   \includegraphics[width=\linewidth]{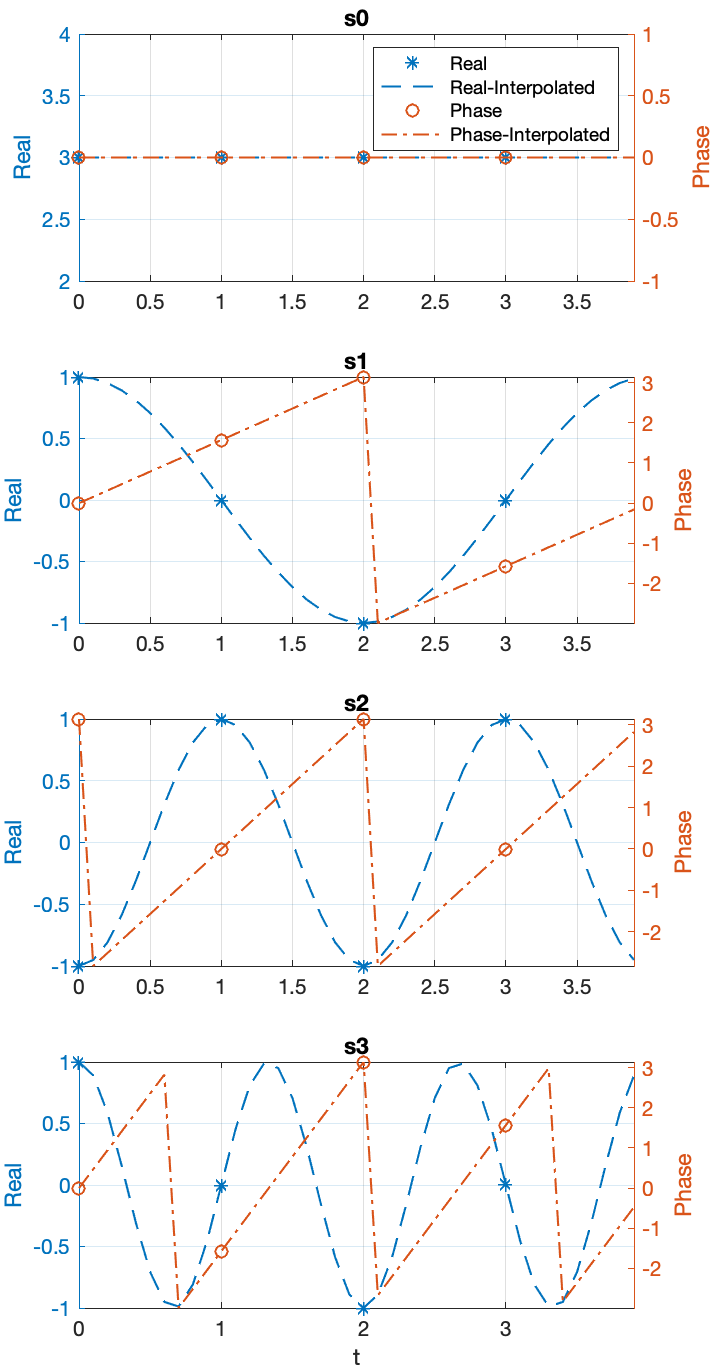}
   \caption{x(t)=``1101"}
   \label{fig:4x1SNN3spikesDelay3}
\end{subfigure}
\centering
\caption{SNN (N=4) Representation of Three Delayed Impulse}
\end{figure*}

{\it Signal Reconstruction Using SNN:} The original signals consisting of three impulses are reconstructed by summing the outputs of four spiking neurons shown in Fig.\ref{fig:4x1SNN3spikesDelay0} to Fig. \ref{fig:4x1SNN3spikesDelay3}. 

\subsubsection{SNN Firing Patterns of Two Delayed Impulses}

Fig.\ref{fig:4x3TwoDelayedImpulse} illustrates the interaction of two impulses. These impulses can be calculated as the superposition of two single impulses discussed earlier. For example, the sequence x(t) = ``1100", is obtained by adding the individual impulse sequences x(t) = `1000"  and x(t) = `0100". It can be observed from the polar plots that the phase pattern of the DFT coefficients of the sequence x(t) = ``1100" is similar to that of the sequence x(t) = ``0100". The former equals the latter added with a the constant coefficient value of the sequence x(t) = ``1000": $Y_k = 1\angle 0, k = 0,1,2,3$. 

The DFT spectrum of the two impulses is obtained by adding the corresponding DFT coefficients $Y_k$. The addition of two vectors in the polar coordinates are shown in the DFT spectrum, and the spiking neurons firing patterns with a constant phase increment $\Delta \phi$ are displayed in the polar plots. Notably, the magnitudes of the four DFT coefficients (Y0, Y1, Y2, and Y3) for the four x(t) signals with two delayed impulses are the same. The DFT spectrum is calculated directly from x(t). Based on the linearity property of the Fourier transform, the DFT coefficients for the combined impulses can also be calculated as the sum of two complex numbers in exponential form (\ref{eq:VectorSum}). 

Table. \ref{tab:SNNPhaseIncrementDelayedImpulse2} lists the DFT coefficients $Y_k$ for the four x(t) signals with two delayed impulses, including their magnitude and phase. The observed distinct sequential firing patterns of the spiking neurons in the SNN suggest the applicability of the phase encoding scheme. However, it should be noted that the earlier phase difference encoding method is not directly suitable for encoding the sum of two complex exponential functions. The interaction between two spiking neurons with different oscillating frequency can produce intriguing patterns based on the phase differences between the two neurons\cite{Zhang22c}, which is explored in ongoing research. 

\begin{equation}
\label{eq:VectorSum}
\begin{aligned}
Y_1 & = r_1 \angle \theta_1 = r_1 e^{i\theta_1}, \quad Y_2 = r_2 \angle \theta_2 = r_1 e^{i\theta_2}\\
Y_s & = Y_1+Y_2 = r_s e^{i\theta_s}\\
\Rightarrow r_s  &= \sqrt {r_1^2 + r_2^2 + 2r_1r_2\cos(\theta_1-\theta_2)}\\
\theta_s &= tan^{-1} \frac{r_1\sin\theta_1 +r_2 \sin\theta_2}{r_1\cos\theta_1 +r_2 \cos\theta_2}
\end{aligned}
\end{equation}

\begin{table}[h]
\caption{SNN Firing Patterns for Two Delayed Impulses: $N=4, \Omega_0 =\pi/2, \omega =\Omega_0 k, k = 0,1,2,3.$}
\centering
\begin{tabular}{|c|c|l|l|l|l|}
\hline
\bf x(t) & $n_0$ & Y0 & Y1  & Y2 & Y3   \\
\hline
1100 & 0 & $2\angle 0$ & $\sqrt 2\angle \frac{-\pi}{4}$ & 0 &  $\sqrt 2\angle \frac{\pi}{4}$   \\
\hline
0110 & 1 & $2\angle 0$ & $\sqrt 2\angle \frac{-5\pi}{4}$  & 0  &  $\sqrt 2\angle \frac{5\pi}{4}$    \\
\hline
0011 & 2 & $2\angle 0$ & $\sqrt 2\angle \frac{5\pi}{4}$  & 0  &  $\sqrt 2\angle \frac{-5\pi}{4}$   \\ 
\hline
1001 & 3 & $2\angle 0$ & $\sqrt 2\angle \frac{\pi}{4}$  & 0  &  $\sqrt 2\angle \frac{-\pi}{4}$   \\
\hline
\end{tabular}
\label{tab:SNNPhaseIncrementDelayedImpulse2}
\end{table}

{\it Signal Reconstruction Using SNN:} The reconstruction of original signals consisting of two impulses can be achieved by summing the outputs of four spiking neurons. Fig.\ref{fig:4x1SNN2spikesDelay0} to Fig.\ref{fig:4x1SNN2spikesDelay3} depict the SNN architecture with four spiking neurons, demonstrating the reconstruction of two impulses with time delays of $n_0 = 0, 1, 2, 3$ respectively. The plots for the real components and phases of the four spiking neurons are interpolated by a factor of 10. The actual samples within one period are indicated by the `*' marker for the real components and the `o' marker for the phase. The original impulse signals can be reconstructed by adding the outputs of these four neurons. Neuron s0 represents a constant value with a frequency of 0. The initial phase of neurons s1, s2, and s3 is determined by the DFT coefficients Y1, Y2, and Y3 respectively, as shown in both the DFT spectrum and the polar plot. The polar plot clearly exhibits a linear phase difference among the four spiking neurons, indicating their sequential firing pattern associated with the original delayed impulse.

\subsubsection{Phase Encoding of Three Delayed Impulses}

For the comparison of the 16 sequences of a 4-point x(t) , Fig.\ref{fig:4x3ThreeDelayedImpulse} shows the x(t) sequences with three delayed impulses, as well as the DFT spectrum and polar plots. And Fig.\ref{fig:4x3FourLeftPatterns} shows the time sequences, DFT spectrum and polar plots for the remaining four x(t) patterns.


%

\section{Conclusion}

In conclusion, this paper presented a novel approach for signal reconstruction using SNN based on Cognitive Informatics and Cognitive Computing principles. By leveraging the DFT, the SNN accurately captured the spectral information of arbitrary time series signals. The encoding of impulse delays and phase differences between frequency components further enhanced the model's capabilities. Experimental results demonstrated the effectiveness of the proposed approach in reconstructing signals with varying time delays. This research contributes to signal processing by providing a biologically inspired SNN framework for accurate signal reconstruction and highlights the potential of cognitive computing in signal analysis.



\bibliographystyle{IEEEtran}
\bibliography{../../bib/references23.7.21}

\end{document}